\documentclass[letterpaper, 10 pt, conference]{ieeeconf}  

\IEEEoverridecommandlockouts                              

\usepackage{graphicx} 
\usepackage{epsfig} 
\usepackage{mathptmx} 
\usepackage{times} 
\usepackage{amsmath} 
\usepackage{amssymb}  
\usepackage{xcolor} 
\usepackage{subcaption}
\usepackage{booktabs}
\usepackage[linesnumbered,ruled,vlined]{algorithm2e}
\usepackage{multirow}
\usepackage[colorlinks=true, citecolor=red]{hyperref}

\title{\LARGE \bf
Event-RGB Adaptive Tracking for Nighttime Highway Perception
}

\author{Haidong Wang$^{1}$, Hengxing Cai$^{1}$, Wanlei Li$^{2}$, Xiaogang Xiong$^{2}$, and Renxin Zhong$^{1,*}$%
\thanks{$^{1}$Haidong Wang, Hengxing Cai, and Renxin Zhong are with the School of Intelligent Systems Engineering, Sun Yat-sen University, Shenzhen, China.
        {\tt\small \{wanghd26,caihx3\}@mail2.sysu.edu.cn, zhrenxin@mail.sysu.edu.cn}}%
\thanks{$^{2}$Wanlei Li and Xiaogang Xiong are with the School of Intelligence Science and Engineering, 
        Harbin Institute of Technology, Shenzhen, China.
        {\tt\small 19b953034@stu.hit.edu.cn, xiongxg@hit.edu.cn}}%
\thanks{$^{*}$Corresponding author: Renxin Zhong.}%
}

\begin{document}

\maketitle
\thispagestyle{empty}
\pagestyle{empty}

\begin{abstract}
Intelligent Transportation Systems deployed on highways predominantly rely on conventional RGB cameras for traffic perception and vehicle tracking. However, highway environments present unique challenges: the absence of artificial lighting infrastructure, combined with high vehicle velocities, results in severely degraded perception performance under low-light conditions. Specifically, nighttime scenarios suffer from motion blur, insufficient exposure, and poor signal-to-noise ratios, which catastrophically impair the reliability of RGB-based sensing systems.
To address these limitations, we propose a novel Joint Event-RGB Adaptive Tracking (JEAT) framework. Unlike existing multi-sensor trackers constrained by rigid, hard-coded prioritization, JEAT merges asynchronous event streams and RGB frames into a unified joint data association optimization. By employing an Adaptive Extended Kalman Filter to continuously estimate measurement noise via NIS statistics, the framework dynamically weights and fuses both modalities, optimally harnessing event streams during dark or high-speed motion while leveraging RGB frames under bright or static conditions.
Furthermore, given the absence of publicly available datasets tailored for event-based highway perception with diverse environmental conditions, we present SEHN, a large-scale synthetic dataset generated using the CARLA simulator. Our dataset encompasses diverse environmental conditions (daytime, nighttime, nighttime with out artificial lighting) and varying traffic densities, providing synchronized RGB imagery and event streams to facilitate multi-modal fusion research. Our code and datasets will be available at \url{https://github.com/haidongwang96/SEHN}.

\end{abstract}


\section{Introduction}
The rapid expansion of highway networks worldwide has driven an increasing demand for robust and reliable Intelligent Transportation Systems (ITS) capable of continuous traffic monitoring, incident detection, and vehicle tracking. Contemporary ITS infrastructures predominantly employ RGB camera-based perception systems, leveraging advances in deep learning for object detection and tracking. By capturing high-resolution visual streams, RGB surveillance cameras enable the precise extraction of multi-dimensional static attributes, such as vehicle models, license plates, and colors, as well as dynamic motion states including instantaneous velocity, traffic volume, and queue length. Through semantic analysis of spatiotemporal sequences, the system further facilitates the automated identification of complex traffic events, such as violations or accidents, and provides quantitative evaluations of the Level of Service for road networks. These high-level data provide essential support for traffic flow prediction, signal control optimization, and early warning systems for anomalous behavior.

\begin{figure}[t!]
\centering
\begin{subfigure}[b]{0.235\textwidth}
\centering
\includegraphics[width=\textwidth]{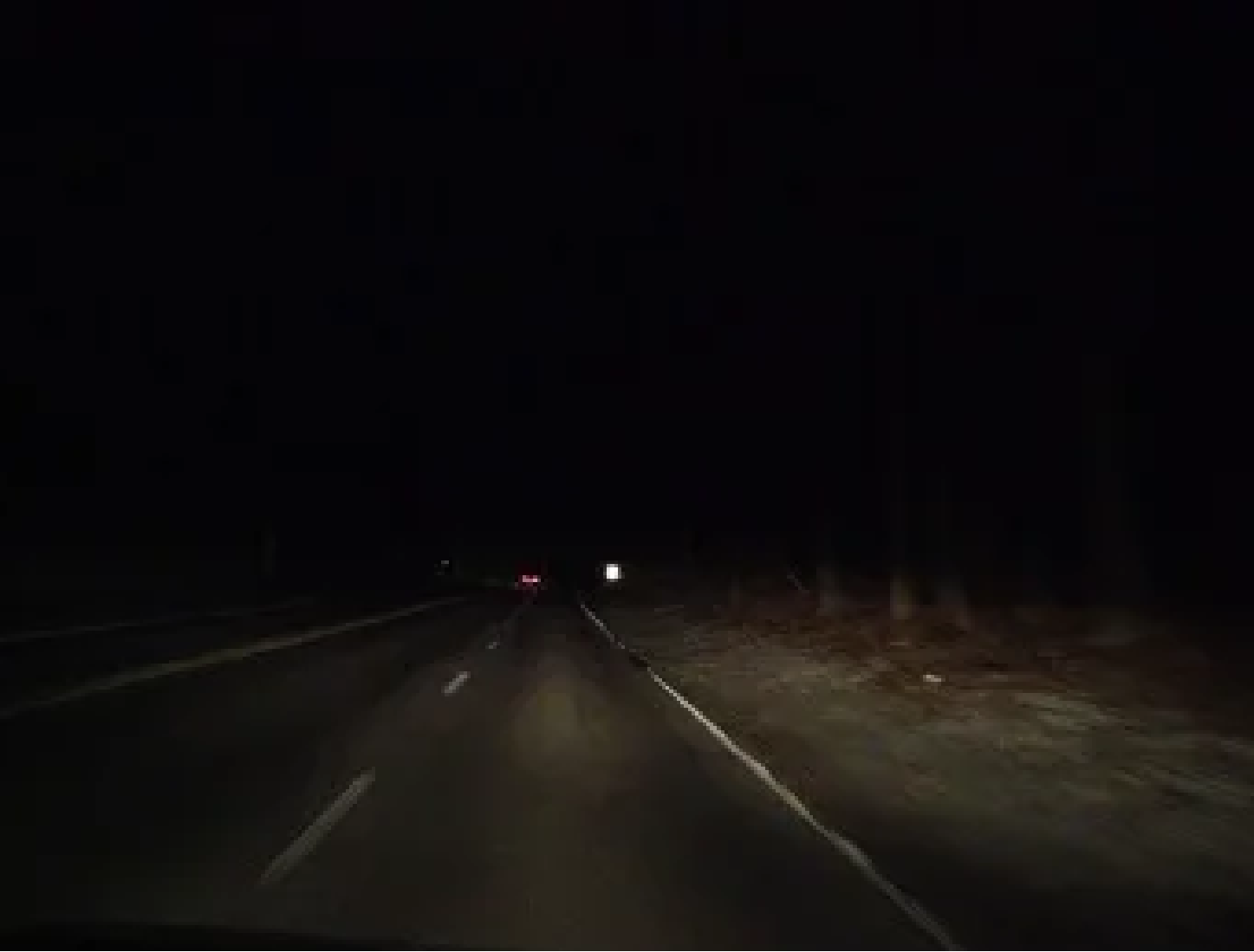}
\caption{Ego motion nighttime highway driving scenario}
\label{fig:ego_motion_nighttime_demo}
\end{subfigure}
\hfill 
\begin{subfigure}[b]{0.235\textwidth}
\centering
\includegraphics[width=\textwidth]{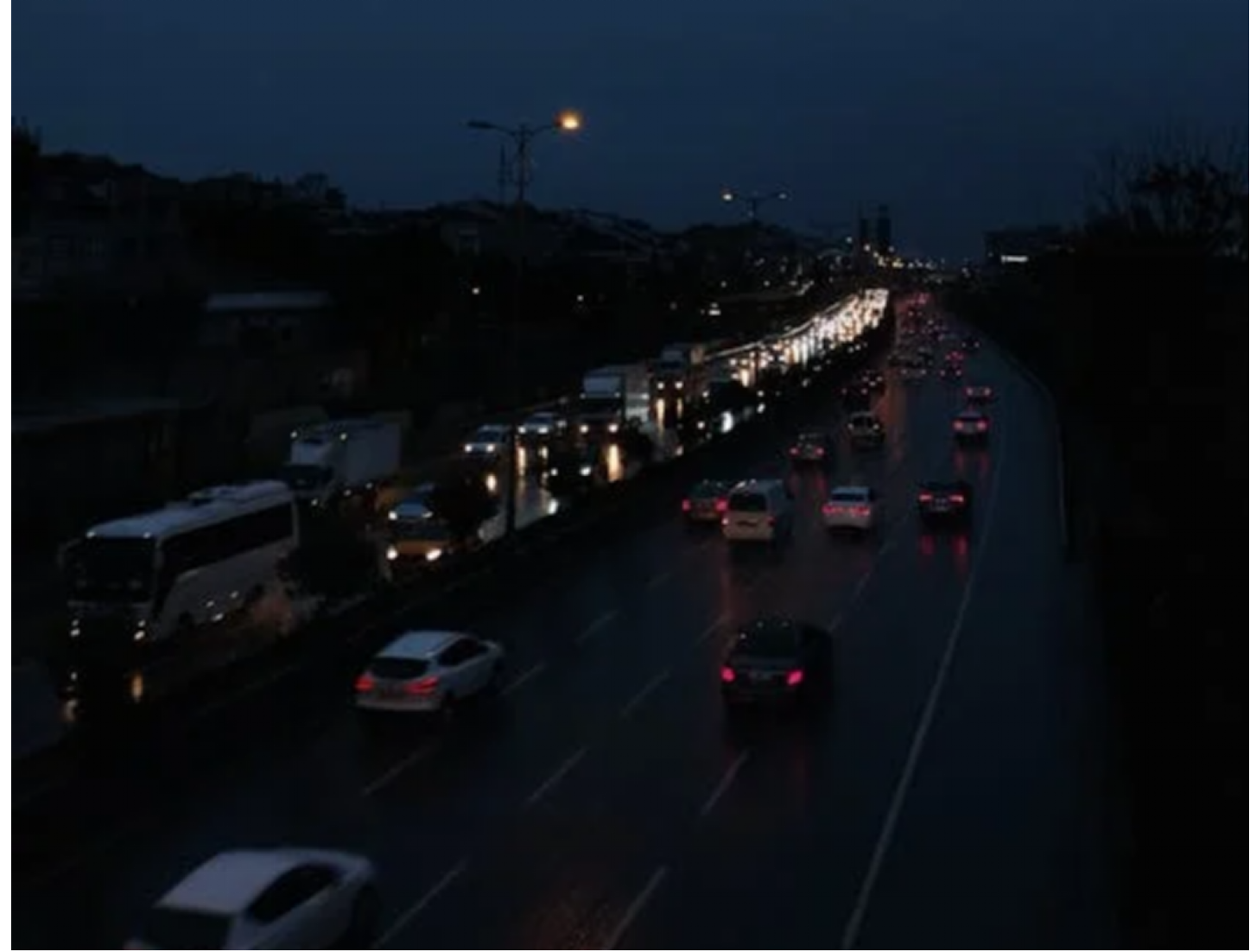}
\caption{Static nighttime highway traffic surveillance scenario}
\label{fig:static_nightime_demo}
\end{subfigure}
\caption{Nighttime highway perception scenarios from (a) ego-motion and (b) static surveillance perspectives. Under low-light conditions with high-speed traffic, conventional RGB cameras suffer from severe motion blur and underexposure, rendering them nearly ineffective for reliable perception}
\label{fig:nightime_demo}
\end{figure}

However, these conventional vision systems have only demonstrated remarkable success under favorable illumination conditions, their performance degrades significantly in challenging scenarios, particularly on highways where artificial lighting is often absent or sparse. Nighttime highway driving presents a particularly hostile environment for conventional vision systems. The combination of high vehicle velocities (typically exceeding 100 km/h) and extremely low ambient illumination leads to severe motion blur, underexposure, and diminished contrast in captured imagery, which is illustrated in Fig~\ref{fig:nightime_demo}. These degradations fundamentally undermine the feature extraction and temporal association mechanisms upon which modern tracking algorithms depend, resulting in catastrophic failure modes including missed detections, identity switches, and fragmented trajectories. Event cameras, also known as Dynamic Vision Sensors (DVS), represent a paradigm shift in visual sensing that offers promising solutions to these challenges. Unlike conventional frame-based cameras that capture synchronous intensity images at fixed intervals, event cameras asynchronously report pixel-level brightness changes with microsecond temporal resolution~\cite{liu2025beyond, gehrig2024low, li2025tumtraf, gehrig2021dsec, detournemire2020large, verma2024etram}.

In the specific context of highway surveillance, traditional RGB-based sensing paradigms encounter significant bottlenecks due to degraded illumination and severe motion blur problems. While other alternative multimodal strategies such as RGB-Thermal or RGB-Depth have been explored, they are often ill-suited for high-speed monitoring; thermal sensors are susceptible to thermal crossover and motion blur, while depth sensors are constrained by a limited sensing distance~\cite{wang2023visevent}.  
Nevertheless, event cameras solely introduce a reciprocal failure mode which events are generated exclusively in response to temporal brightness changes, stationary or slowly moving objects produce little to no signal in the event stream, rendering them effectively invisible to event-only pipelines. On highways, where abnormally stopped vehicles constitute one of the most safety-critical hazards, this blind spot is unacceptable. The failure modes of the two modalities are therefore complementary rather than overlapping—RGB collapses under low-light, high-speed conditions while event sensing fails in the presence of static targets—giving rise to a cross-modal failure problem that demands an adaptive fusion mechanism capable of dynamically shifting trust to whichever sensor remains operational.
Therefore the integration of RGB and event-based sensors presents a biologically inspired neuromorphic solution. By leveraging asynchronous pulse delivery, this modality provides microsecond temporal resolution that effectively eliminates motion blur, a critical requirement for third-person surveillance of high-velocity vehicles. 


The complementarity between the two sensing paradigms is rooted in their fundamentally different mechanisms. Event cameras respond to temporal brightness changes with microsecond latency, making them inherently resilient to the motion blur and underexposure that cripple frame-based sensors at night; however, they are blind to objects that remain stationary in the scene. RGB cameras, conversely, capture full-frame intensity snapshots that encode appearance and texture regardless of object motion, yet their fixed integration time renders them vulnerable to the very high-speed, low-light regimes in which event sensors thrive. These opposing failure characteristics suggest that neither modality should be permanently prioritized; instead, the optimal sensor to trust varies continuously with the instantaneous illumination and motion profile. We therefore propose the Joint Event-RGB Adaptive Tracking (JEAT) framework, which departs from conventional fixed-priority fusion by formulating a single joint data association problem over all detections from both sensors. To realize dynamic modality reweighting without manual intervention, JEAT incorporates an Adaptive Extended Kalman Filter whose measurement noise covariance is continuously recalibrated through the Normalized Innovation Squared (NIS) statistic, allowing the cost of each detection to reflect the sensor's instantaneous reliability. Existing event camera datasets are predominantly captured from ego-motion perspectives under favorable illumination, leaving a critical gap for static-view highway surveillance in extreme low-light conditions. No publicly available dataset, to our knowledge, provides co-registered RGB and event streams on highways devoid of artificial lighting where vehicles travel only with headlights in the dark scene. We introduce SEHN (Synthetic Event-based Highway Nighttime), a large-scale dataset generated through a highly configurable CARLA-based simulation pipeline that spans the full illumination spectrum with controlled traffic density and speed distributions, providing a needed dataset for evaluating multi-modal fusion algorithms under extreme traffic scenarios.

Contributions of this work:
\begin{itemize}
    \item We propose the Joint Event-RGB Adaptive Tracking (JEAT) framework, a principled multi-sensor fusion approach that unifies asynchronous RGB and event camera detections within a single data association optimization targeting on highway vehicle tracking. 
    \item We present SEHN, a large-scale synthetic multimodal dataset built on CARLA, specifically targeting nighttime highway perception with synchronized RGB and event streams across diverse illumination and traffic conditions.
    \item We demonstrate that JEAT effectively resolves cross-modal failure by performing adaptive fusion at the tracking level that event sensors compensate for RGB degradation in dark, high-speed scenarios, while RGB data covers the inherent blindness of event sensors to stationary objects. 
\end{itemize}

\section{Related Work}

\subsection{Event-based Detection and Tracking}
The foundation of robust event-based tracking lies in accurate feature or object detection. Previous works in pure event-based detection typically transform sparse, asynchronous event streams into event representations~\cite{gallego2020event}, such as  event frames, time-surfaces, voxel grids. 
Event-based detection has also witnessed rapid advancements in recent years,~\cite{perot2020learning} directly employs 2D histogram frames and leverages frame-based detection methods. To address the inherent sparsity of event data, approaches~\cite{gehrig2024low, gehrig2022pushing, schaefer2022aegnn} based on graph neural networks exploit sparse connectivity to significantly reduce computational complexity. Meanwhile, transformer-based approaches~\cite{peng2024scene, gehrig2023recurrent} have improved detection accuracy by capturing spatiotemporal dependencies through memory mechanisms.

Early event-based tracking methods relied on motion compensation and mean-shift clustering on spatiotemporal event clouds~\cite{mitrokhin2018event}. With the advent of deep learning, detection-based approaches evolved rapidly. 
The broader multi-object tracking literature has been dominated by the tracking-by-detection framework, which decouples the detection and association stages to achieve a favorable balance between accuracy and computational efficiency. Representative lightweight trackers in this category include SORT~\cite{bewley2016simple}, Bytetrack~\cite{zhang2022bytetrack}, BoT-SORT~\cite{aharon2022bot}. In the specific context of traffic surveillance, TUMTraf EMOT~\cite{li2025tumtraf} extends the tracking-by-detection paradigm.

\subsection{Event-based Dataset}
High quality Event datasets serves as a fundamental prerequisite for addressing complex vision tasks. Nevertheless, event-based datasets remain limited in scale and diversity compared to established RGB collections. This scarcity is particularly pronounced in challenging scenarios such as third-person traffic surveillance under low-light conditions. The physical acquisition of such data faces significant barriers due to the limited hardware accessibility and technical maturity of event sensors. Current methodologies for generating event datasets primarily encompass two paradigms, including direct acquisition, event synthesis. Beyond these acquisition paradigms, event recordings can be further categorized by their observation perspectives into ego-motion-based captures and static view scenarios. While the former typically involves sensors mounted on moving platforms for environmental mapping, the latter focuses on fixed-viewpoint monitoring, which is critical for long-term surveillance but remains relatively under-explored.
The majority of direct acquisition event-based datasets focus on ego-motion scenarios, with representative works including DSEC~\cite{gehrig2021dsec}, MVSEC~\cite{zhu2018multivehicle}, 1MegaPixel~\cite{perot2020learning}, and Gen 1~\cite{detournemire2020large}. Datasets dedicated to static-view settings remain relatively scarce, with ETram~\cite{verma2024etram} and TUMTraf event~\cite{cress2024tumtraf} being notable examples.  
Event synthesis methodologies diverge into video-to-event conversion and simulator-based synthesis. Datasets utilizing video-to-event conversion include RPG DAVIS~\cite{mueggler2017event}, ESIM~\cite{rebecq2018esim}, and v2e~\cite{hu2021v2e}, whereas simulator-based synthesis encompasses datasets such as SEVD~\cite{aliminati2024sevd} for ego-motion and SeTram~\cite{tan2025real} for static views. Despite these advancements, to our knowledge, no existing dataset specifically targets nighttime highway perception under extreme conditions. This deficiency is particularly critical in scenarios involving ultra-low-light environments without street lighting, where the combination of minimal illumination and high-speed traffic significantly degrades the information richness of the RGB modality. Consequently, the role of event cameras becomes indispensable for reliable perception in such challenging settings.

\section{Method}


\subsection{Problem Formulation}

Our proposed method receives detections from two asynchronous sensors: RGB camera and event camera.  The RGB camera produces standard intensity frames, while the event camera outputs an asynchronous stream of events $\mathcal{E}=\{e_i\}$, where each event $e_i=(x_i,\,y_i,\,t_i,\,p_i)$ records the pixel coordinate $(x_i,y_i)$, timestamp $t_i$ with microsecond resolution, and polarity $p_i\in\{-1,+1\}$ indicating brightness decrease or increase. Object detections are obtained by off-the-shelf detectors for each modality (e.g., YOLOv11~\cite{khanam2024yolov11} for RGB frames and an RVT \cite{gehrig2023recurrent} for event streams); the detector design is outside the scope of this work. Let $\{t_1^{\mathrm{rgb}},t_2^{\mathrm{rgb}},\dots\}$ and $\{t_1^{\mathrm{evt}},t_2^{\mathrm{evt}},\dots\}$ denote the respective detection timestamps; these two sequences are interleaved and processed in temporal order without explicit synchronization. The goal of multi-object tracking (MOT) is to estimate the set of vehicle trajectories $\mathcal{T}=\{T_1,\dots,T_N\}$ from this asynchronous stream. Each trajectory $T_i$ is a time-indexed sequence of states:
\begin{equation}
    T_i = \bigl\{(\mathbf{x}_i^{(t)},\; t) \;\big|\; t\in[t_i^{\mathrm{start}},\; t_i^{\mathrm{end}}]\bigr\},
\end{equation}
where $\mathbf{x}_i^{(t)}\in\mathbb{R}^6$ encodes the position, velocity, and acceleration of the $i$-th vehicle at time $t$. At each timestamp $t$, the detectors produce a detection set $\mathcal{D}_t=\{d_1,\dots,d_{M_t}\}$, with each detection $d_j=(\mathbf{z}_j,\,R_j,\,s_j,\,p_j)$ comprising the 2-D measurement $\mathbf{z}_j\in\mathbb{R}^2$, noise covariance $R_j$, sensor label $s_j\in\{\mathrm{rgb},\,\mathrm{evt}\}$, and confidence score $p_j\in[0,1]$.

\begin{figure}[thbp] 
    \centering 
    \includegraphics[width=\linewidth]{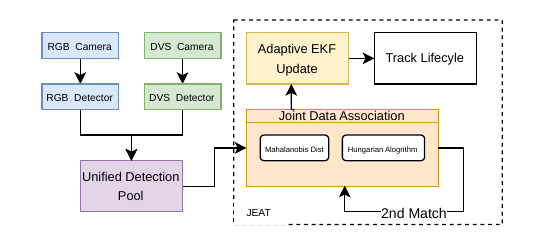} 
    \caption{The architecture of our JEAT framework}
    \label{fig:method} 
\end{figure}

\subsection{Motion and Measurement Models}
\label{sec:motion}

We adopt a constant-acceleration (CA) kinematic model to describe vehicle motion in the image plane. The 6-D state vector is $\mathbf{x} = [x,\, y,\, v_x,\, v_y,\, a_x,\, a_y]^\top \in\mathbb{R}^6$, where $(x,y)$ denotes the centroid position, $(v_x,v_y)$ the velocity, and $(a_x,a_y)$ the acceleration. Given an arbitrary time interval $\Delta t$ between consecutive sensor events, the discrete-time state transition and covariance prediction are
\begin{equation}
    \begin{aligned}
        \mathbf{x}_{k+1} &= F(\Delta t)\,\mathbf{x}_k + \mathbf{w}_k \\
        P_{k|k-1} &= F(\Delta t)\,P_{k-1|k-1}\,F(\Delta t)^\top + Q(\Delta t)
    \end{aligned}
    \label{eq:state_pred}
\end{equation}
where $\mathbf{w}_k\sim\mathcal{N}(\mathbf{0},Q)$ is the process noise. The $x$- and $y$-axes are fully decoupled, so $F(\Delta t)$ and $Q(\Delta t)$ each factor into a Kronecker product with $I_2$. The transition matrix encodes the standard kinematic relations (position integrates velocity and acceleration; velocity integrates acceleration; acceleration is assumed constant). The process noise covariance $Q$ is derived by modeling jerk as continuous white noise with spectral density $\sigma_j$. Notably, the position-uncertainty term in $Q$ scales as $\mathcal{O}(\Delta t^5)$, implying that predicted uncertainty grows rapidly with the prediction horizon. This gives the higher-rate event camera ($\Delta t\!\approx\!10$\,ms) a natural advantage over the RGB camera ($\Delta t\!\approx\!50$\,ms) in maintaining a tight $P_{k|k-1}$ and thus more reliable data association.

Both modalities observe the 2-D centroid position. The measurement equation is
\begin{equation}
\begin{aligned}
        \mathbf{z}_k &= H\,\mathbf{x}_k + \mathbf{v}_k, \\
    H &= \begin{bmatrix} I_{2\times 2} & \mathbf{0}_{2\times 4} \end{bmatrix}
\end{aligned}
\end{equation}
where $\mathbf{v}_k\sim\mathcal{N}(\mathbf{0},R_k)$. The sensor-specific noise covariance is modeled as $R_k = R_{\mathrm{base}}^{(s)}\cdot f_{\mathrm{adapt}}(\cdot)\cdot g(p_k)$ with $g(p_k)=1/\max(p_k,0.1)$, where $R_{\mathrm{base}}^{(s)}$ is the per-sensor prior, $f_{\mathrm{adapt}}$ is the online adaptive factor (Section~\ref{sec:adaptive}), and $g(p_k)$ down-weights low-confidence detections. In the joint association framework, $R_k^{(s)}$ directly enters the innovation covariance, so a sensor with smaller $R$ produces a smaller Mahalanobis cost and is preferentially selected by the Hungarian algorithm without explicit priority encoding.

\subsection{State Update}
\label{sec:ekf}

State estimation follows an Adaptive Extended Kalman Filter (AEKF) cycle, which augments the standard EKF with online measurement noise adaptation (Section~\ref{sec:adaptive}). After predicting all tracks to the current timestamp via Eq.~(\ref{eq:state_pred}), each matched track incorporates the associated measurement $\mathbf{z}_k$ by computing the innovation $\mathbf{y}_k = \mathbf{z}_k - H\hat{\mathbf{x}}_{k|k-1}$, the innovation covariance $S_k = HP_{k|k-1}H^\top + R_k^{(s)}$, and the Kalman gain $K_k = P_{k|k-1}H^\top S_k^{-1}$, yielding the posterior:
\begin{equation}
    \begin{aligned}
        &\hat{\mathbf{x}}_{k|k} = \hat{\mathbf{x}}_{k|k-1} + K_k\,\mathbf{y}_k \\
        &P_{k|k} = (I - K_k H)\,P_{k|k-1}\,(I - K_k H)^\top + K_k\,R_k\,K_k^\top
    \end{aligned}
    \label{eq:state_update}
\end{equation}
where the covariance update adopts the Joseph form to preserve positive definiteness. When a track receives detections from both sensors at the same timestamp via the second-pass mechanism (Section~\ref{sec:association}), two successive updates are applied sequentially, each using the sensor-specific $R_k^{(s)}$. This is mathematically equivalent to a single batch update with a stacked measurement vector and block-diagonal noise covariance, but avoids constructing the augmented system.

\subsection{Joint Data Association}
\label{sec:association}

Existing multi-sensor trackers typically employ a fixed-priority sequential processing framework. This rigid hierarchy restricts subsequent search spaces based on initial matches, while the inflexible, hard-coded prioritization fails to accommodate dynamic environmental changes.
Instead, we merge all detections into a single optimization. At each timestamp $t$, the $M_{\mathrm{rgb}}$ detections from the RGB detector and the $M_{\mathrm{evt}}$ detections from the event detector are combined into a unified set ${D}_t = {D}_t^{\mathrm{rgb}}\cup{D}_t^{\mathrm{evt}}$ of $M=M_{\mathrm{rgb}}+M_{\mathrm{evt}}$ detections, each retaining its sensor label $s_j$. For track $T_i$ and detection $d_j$ from sensor $s_j$, the sensor-adaptive Mahalanobis distance is
\begin{equation}
    \begin{aligned}
         &d^2(T_i,d_j) = \mathbf{y}_{ij}^\top S_{ij}^{-1}\,\mathbf{y}_{ij} \\
        &S_{ij} = HP_iH^\top + R_k^{(s_j)}
    \end{aligned}
    \label{eq:mahal}
\end{equation}
where $\mathbf{y}_{ij} = \mathbf{z}_j - H\hat{\mathbf{x}}_i$. Since $R_k^{(s_j)}$ is sensor-specific, the same residual yields different costs for different sensors, providing an implicit competition mechanism. Entries exceeding a chi-square gate are set to $+\infty$, and the Hungarian algorithm solves the resulting $N\!\times\!M$ cost matrix, producing matched pairs $\mathcal{M}$, unmatched tracks $\mathcal{U}_\mathcal{T}$, and unmatched detections $\mathcal{U}_\mathcal{D}$.

Since each track receives at most one match, a target observed by both sensors only gets its lower-cost detection assigned. An optional second pass therefore searches $\mathcal{U}_\mathcal{D}$ for complementary detections: for each $(i,j_1)\in\mathcal{M}$, the closest gated detection from the other modality triggers a second AEKF update (Section~\ref{sec:ekf}) and is removed from $\mathcal{U}_\mathcal{D}$. The joint formulation is provably no worse than sequential association, as the sequential solution is a feasible---but generally suboptimal---solution to the joint problem.

\begin{algorithm}[htbp]
\caption{Joint RGB-Event adaptive Tracking}\label{alg:main}
\KwIn{Detection stream $\{(\mathbf{D}_t,\,t)\}$ in temporal order; track set $\mathbf{T}\!\leftarrow\!\emptyset$}
\KwOut{Track trajectories $\mathbf{T}$}

\For{each detection batch $(\mathbf{D}_t,\,t)$}{
    \For{each track $T_i\in \mathbf{T}$}{
        $\Delta t \leftarrow t - T_i.\mathrm{last\_time}$\;
        \If{$\Delta t > 0$}{
            $T_i.\textsc{Predict}(\Delta t)$ 
        }
    }
    $\mathbf{D}_{\mathrm{all}} \leftarrow \mathbf{D}_t^{\mathrm{rgb}} \cup \mathbf{D}_t^{\mathrm{evt}}$\;
    $(\mathbf{M},\,\mathbf{U}_{\mathbf{T}},\,\mathbf{U}_{\mathbf{D}}) \leftarrow \textsc{JointAssociate}(\mathbf{D}_{\mathrm{all}},\,\mathbf{T})$\;
    
    \For{$(i,\,j)\in \mathbf{M}$}{
        $T_i.\textsc{Update}(\mathbf{z}_j,\;R^{(s_j)},\;t)$\;
    }
    
    \For{$(i,\,j_1)\in \mathbf{M}$}{
        $s_2 \leftarrow \mathrm{opposite}(s_{j_1})$\;
        $\mathbf{C} \leftarrow \{d\!\in\!\mathbf{U}_{\mathbf{D}} : s_d\!=\!s_2 \;\wedge\; d^2(T_i,d)\le\chi^2_{\mathrm{gate}}\}$\;
        \If{$\mathbf{C}\neq\emptyset$}{
            $j_2\leftarrow\arg\min_{d\in \mathbf{C}}d^2(T_i,d)$\;
            $T_i.\textsc{Update}(\mathbf{z}_{j_2},\;R^{(s_2)},\;t)$\;
            $\mathbf{U}_{\mathbf{D}}\leftarrow \mathbf{U}_{\mathbf{D}}\setminus\{d_{j_2}\}$\;
        }
    }
    
    \For{$d\in \mathbf{U}_{\mathbf{D}}$}{
        $\mathbf{T}\leftarrow \mathbf{T}\cup\{\textsc{CreateTrack}(d,t)\}$\;
    }
    
    \For{$T_i\in \mathbf{U}_{\mathbf{T}}$}{
        $T_i.\textsc{MarkMissed}(t)$\;
        \If{$T_i.\textsc{ShouldDelete}()$}{
            $\mathbf{T}\leftarrow \mathbf{T}\setminus\{T_i\}$\;
        }
    }
}
\KwRet $\mathbf{T}$\;
\end{algorithm}

\subsection{Adaptive Measurement Noise Estimation}
\label{sec:adaptive}

Accurate estimation of $R_k^{(s)}$ is critical because it directly determines each sensor's competitiveness in the joint cost matrix. We maintain a separate scalar scaling factor $s_k$ per sensor per track, adapting $R$ online via the Normalized Innovation Squared (NIS) statistic.

For a consistent filter, the NIS $\varepsilon_k = \mathbf{y}_k^\top S_k^{-1}\mathbf{y}_k$ follows a $\chi^2(n_z)$ distribution with $\mathbb{E}[\varepsilon_k]=n_z=2$. Deviations from this expectation indicate a mismatched $R$. The adapted covariance is $R_k^{\mathrm{adapted}}=s_k\cdot R_0$, where $s_k$ is updated at each step via
\begin{equation}
    s_{k+1} = \max\!\bigl(s_{\min},\;\min\!\bigl(s_k(1+\eta(\varepsilon_k / n_z-1)),\;s_{\max}\bigr)\bigr) 
    \label{eq:nis}
\end{equation}
with adaptation rate $\eta=0.1$ and clamping bounds $s_{\min}=0.5$, $s_{\max}=5.0$ that prevent the scaling factor from diverging. When $\varepsilon_k>n_z$ the residuals exceed expectation and $s_k$ increases; when $\varepsilon_k<n_z$ the filter is over-conservative and $s_k$ decreases. Detection confidence is further incorporated as $R_k^{\mathrm{final}} = R_k^{\mathrm{adapted}}/\max(p_k,0.1)$, inflating the noise for low-confidence detections. The continuously adapted $R_k^{(s)}$ enters the Mahalanobis distance (Eq.~\ref{eq:mahal}) and causes the sensor preference to shift dynamically---in high-speed scenarios the event camera's frequent updates yield smaller $d_{\mathrm{evt}}^2$, while in static conditions the RGB sensor dominates---eliminating manual priority tuning.

\subsection{Track Lifecycle Management}
\label{sec:lifecycle}

Each track progresses through three states: \textsc{Tentative}, \textsc{Confirmed}, and \textsc{Deleted}. Unmatched detections in $\mathcal{U}_\mathcal{D}$ spawn new \textsc{Tentative} tracks. A tentative track is promoted to \textsc{Confirmed} once it satisfies
\begin{equation}
    \tau_{\mathrm{age}} \ge \tau_{\mathrm{confirm}} \;\;\wedge\;\; n_{\mathrm{hits}} \ge n_{\min},
\end{equation}
where $\tau_{\mathrm{age}}=t_{\mathrm{current}}-t_{\mathrm{creation}}$ is the track age, $\tau_{\mathrm{confirm}}=0.5$\,s is the minimum confirmation age, $n_{\mathrm{hits}}$ counts successful associations, and $n_{\min}=5$. A track is transitioned to \textsc{Deleted} when
\begin{equation}
    t_{\mathrm{current}} - t_{\mathrm{last\_update}} > \tau_{\max},
\end{equation}
with $\tau_{\max}=0.5$\,s for confirmed tracks (allowing tolerance for temporary occlusions) and $\tau_{\max}=0.2$\,s for tentative tracks (quickly removing false positives). Only \textsc{Confirmed} tracks are included in the final output trajectories. The overall algorithm is summarized in Algorithm~\ref{alg:main}.

\section{Dataset Overview}

\begin{figure*}[ht]
\centering
\subfloat[Street lights on, before BA noise mitigation]{\includegraphics[width=\columnwidth]{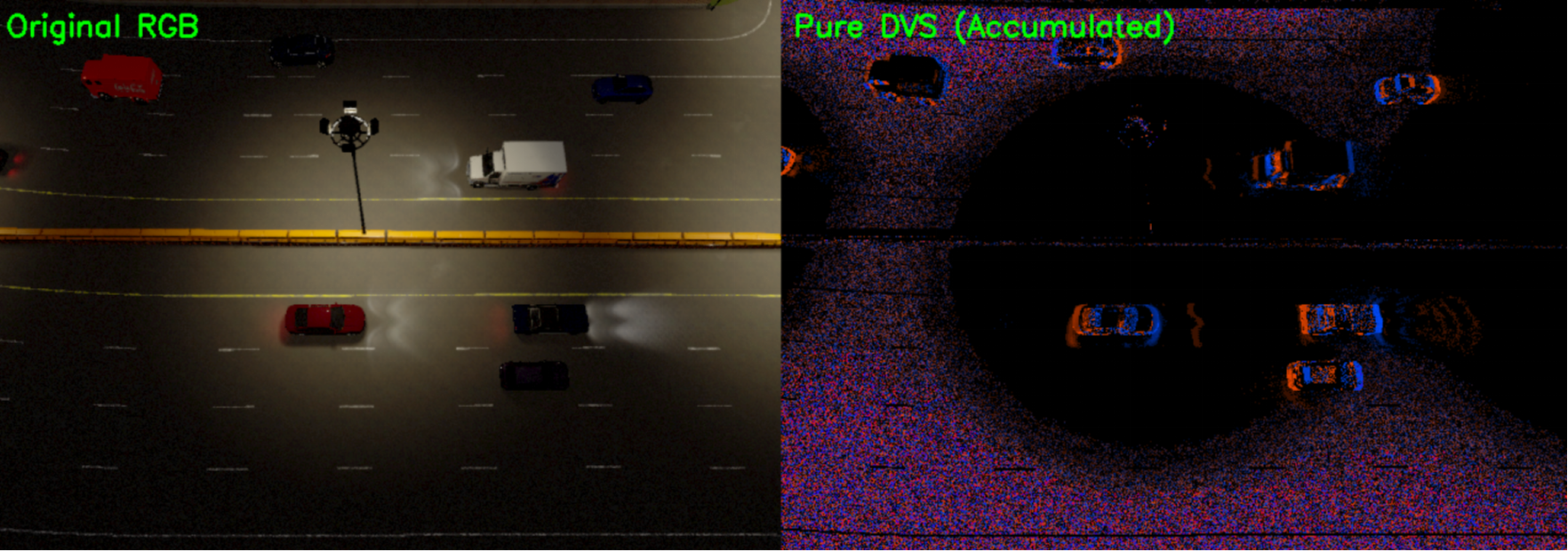}\label{fig:light_on_before}}
\hfill
\subfloat[Street lights off, before BA noise mitigation]{\includegraphics[width=\columnwidth]{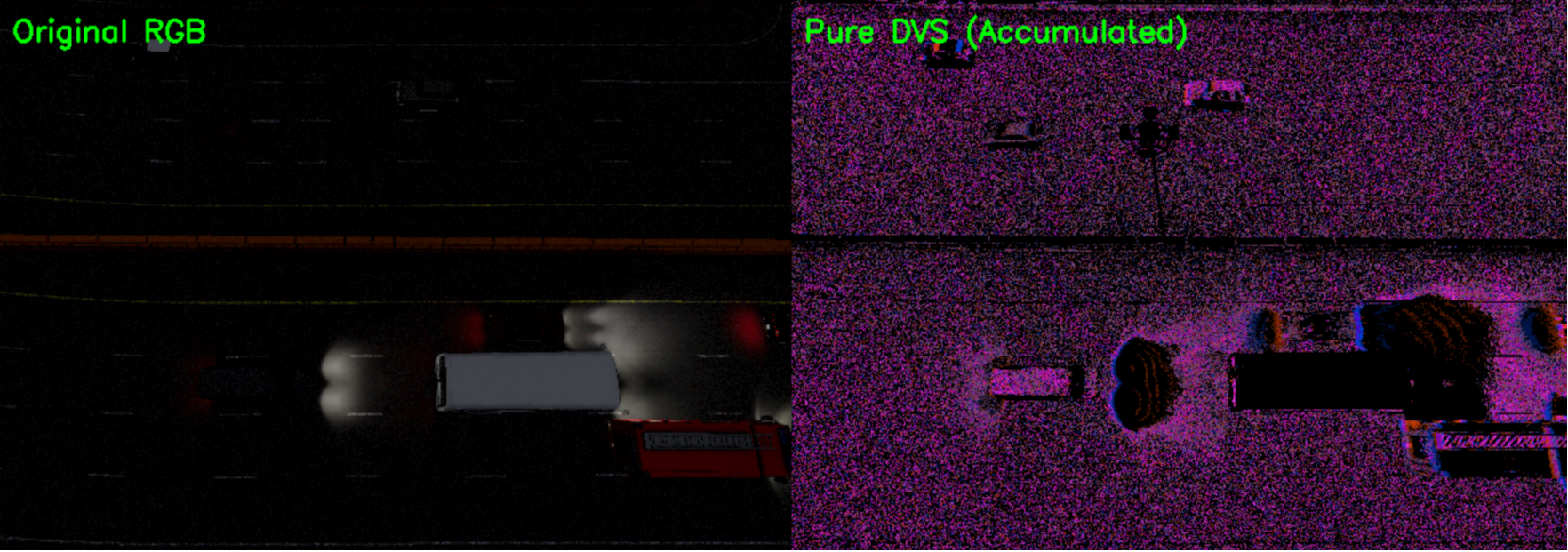}\label{fig:light_off_before}}\\[2pt]
\subfloat[Street lights on, after BA noise mitigation]{\includegraphics[width=\columnwidth]{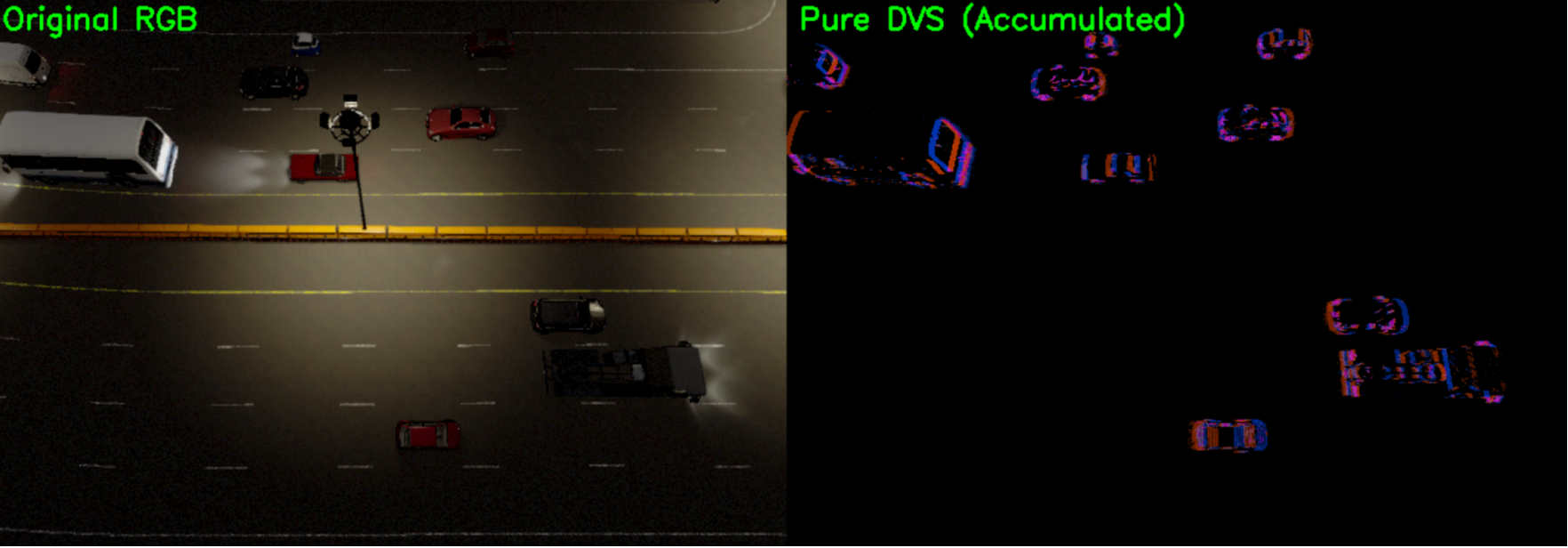}\label{fig:light_on_after}}
\hfill
\subfloat[Street lights off, after BA noise mitigation]{\includegraphics[width=\columnwidth]{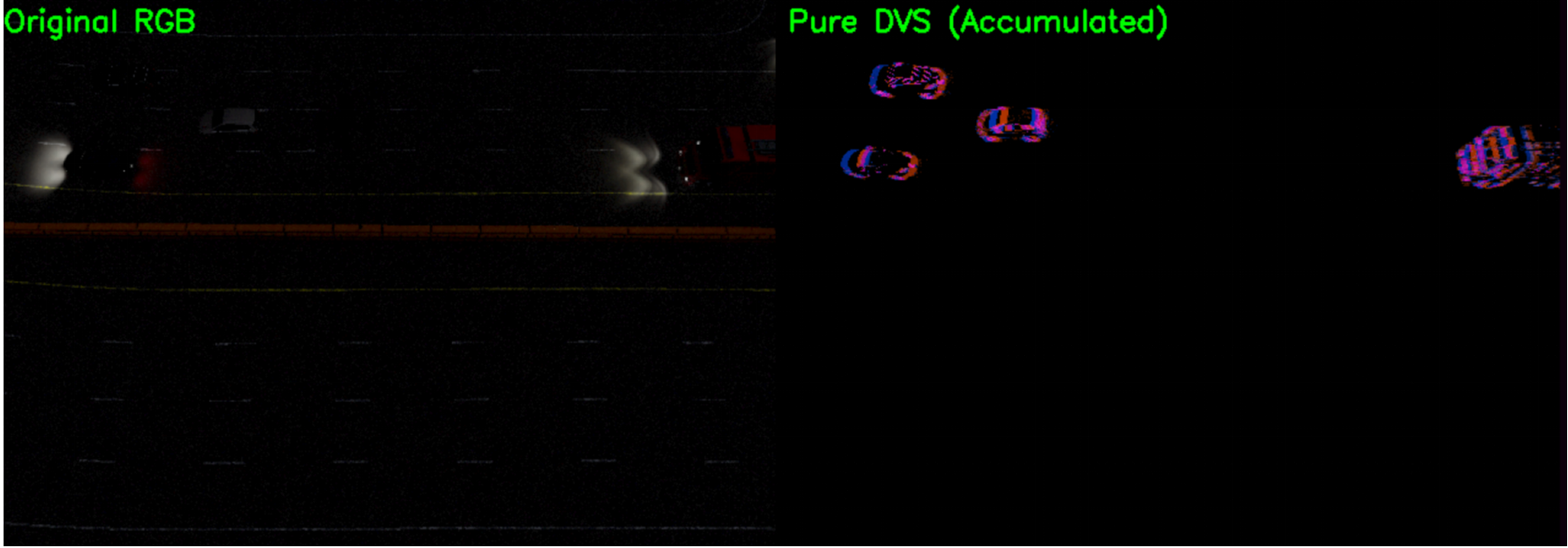}\label{fig:light_off_after}}
\caption{RGB and DVS event frames before (top) and after (bottom) noise mitigation under two lighting conditions: with street lights (left) and without street lights (right). The proposed parameter optimization significantly reduces background noise while preserving motion-correlated events.}
\label{fig:dvs_denoise}
\end{figure*}

\subsection{Dataset setup}

\begin{figure*}[htbp] 
    \centering 
    \includegraphics[width=0.9\textwidth]{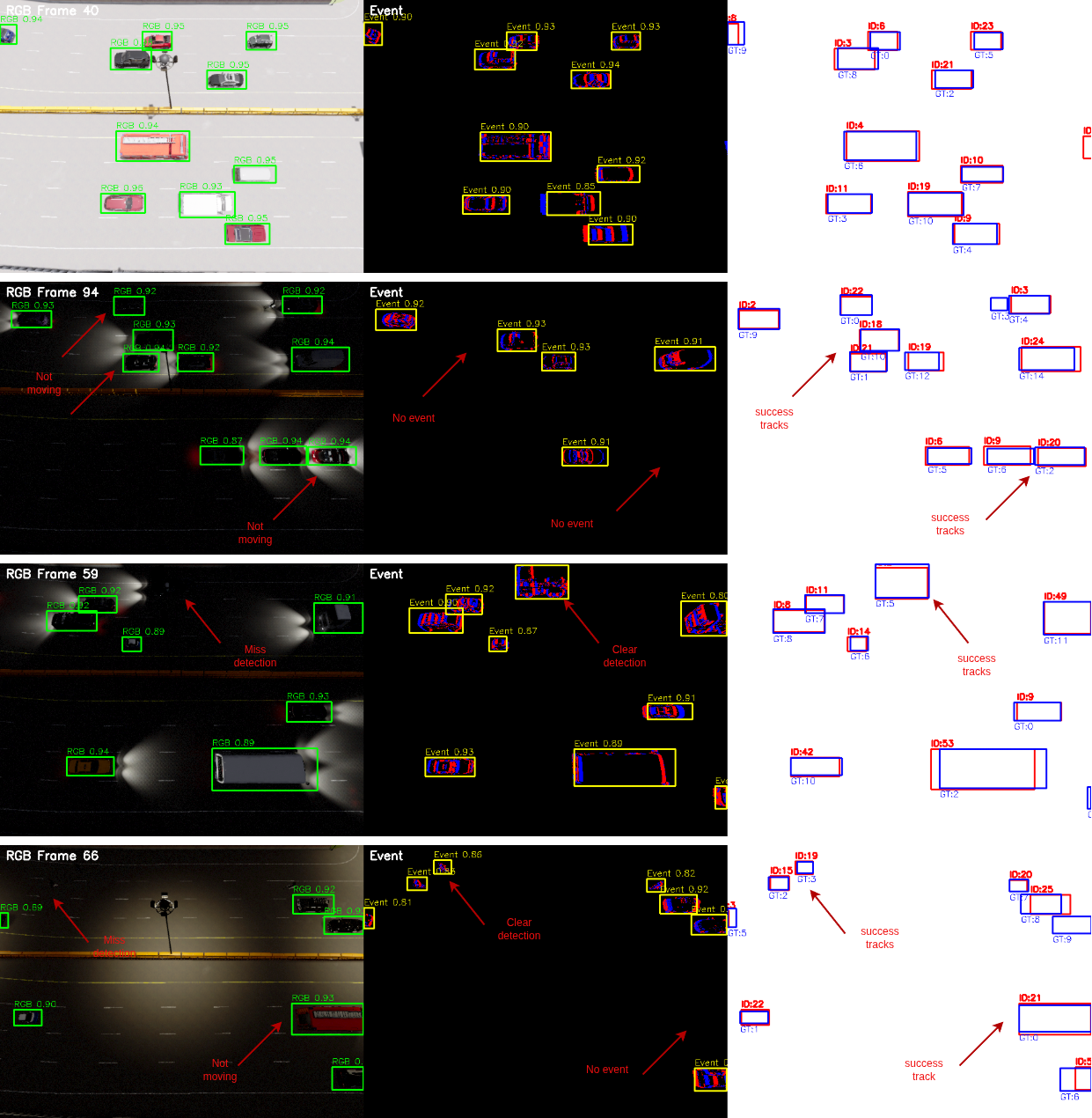} 
    \caption{Object Detection Performance of RGB and Event Modalities on SEHN Dataset. (1) Column-wise: From left to right, the columns present the RGB images with their corresponding detection results, the event frames (50ms interval) along with their detection results, and the tracking visualization, blue boxes denote the ground truth trajectories, while red boxes represent the results produced by the proposed method. (2) Row-wise: The first row presents the outcomes for daytime highway traffic scenarios. The second row depicts a nighttime highway environment without streetlights, where an abnormal parking event results in a lack of event data. The third and fourth rows demonstrate instances of RGB perception failure in nighttime highway settings, comparing environments without and with streetlights, respectively.}
    \label{fig:detection_performance} 
\end{figure*}

We developed a configurable simulation pipeline based on the CARLA simulator to generate DVS-RGB multimodal datasets for nighttime vehicle detection and tracking, SEHN(\textbf{S}ynthetic \textbf{E}vent-based \textbf{H}ighway \textbf{N}ighttime dataset). Unlike existing event camera datasets constrained to fixed recording conditions, our framework enables systematic data generation across diverse environmental and traffic scenarios.
We employ the CARLA open-source driving simulator (version 0.9), which is built upon the Unreal Engine 4 rendering pipeline. The simulation scene is constructed in the Town04 map, which features a multi-lane highway layout suitable for vehicle detection research. The environment is configured to represent different lighting conditions by setting the solar altitude with optional disabling of roadside street lighting to simulate extreme low-light scenarios. Traffic participants—comprising sedans, SUVs, trucks, and vans—are spawned via the CARLA Traffic Manager with autopilot enabled. To maximize scene diversity and vehicle density variation across the dataset, we adopt a segmented collection protocol. The full recording session is divided into 500 segments with 100 frames each. At the beginning of each segment, a new set of vehicles is spawned at positions within $d_\text{spawn} = 150$ m of the camera forward direction. Ground-truth 2D bounding boxes are computed via a full 3D-to-2D projection pipeline. For each vehicle present in the scene, the eight vertices of its 3D axis-aligned bounding box (in local object coordinates) are transformed to world coordinates using the vehicle's rigid-body, and subsequently projected onto the image plane through the camera's extrinsic matrix and intrinsic matrix, as shown in Fig.~\ref{fig:detection_performance}.

\subsection{DVS Noise Mitigation Under Extreme Low-Light Conditions}

A critical challenge during dataset generation is the substantial background activity (BA) noise produced by the CARLA DVS sensor under ultra-low-light conditions, as illustrated in Fig.~\ref{fig:dvs_denoise} where the unlit highway scenario exhibits dense spurious events that obscure genuine motion signals. 
This reflects the physical behavior of real event cameras: below approximately 10~lux, photoreceptor bandwidth contracts proportionally with illumination, causing temporal noise accumulation that manifests as spurious events~\cite{gracca2023shining}. In CARLA's ESIM-based implementation, each pixel triggers an event when the log-intensity change exceeds a contrast threshold $C$:
\begin{equation}
\left|\log\!\left(\epsilon + \frac{I_t}{255}\right) - \log\!\left(\epsilon + \frac{I_{\text{ref}}}{255}\right)\right| > C,
\label{eq:dvs_trigger}
\end{equation}
where $\epsilon$ is a small constant preventing numerical singularity and $I_{\text{ref}}$ is the stored reference intensity. Under low light, intensity values cluster near zero and the logarithmic mapping amplifies minute fluctuations into threshold-crossing events indistinguishable from genuine motion signals.

To suppress this noise while preserving physically meaningful events, we apply three targeted parameter modifications. First, we set the threshold noise parameters $\sigma_{C^+}$ and $\sigma_{C^-}$ to zero, disabling the Gaussian perturbation that CARLA uses to model inter-pixel threshold mismatch. Under this perturbation, the effective threshold is sampled as
$C_{\text{eff}} = C + \mathcal{N}(0,\,\sigma_C).$
While non-zero $\sigma_C$ improves realism under adequate illumination, in dark scenes it stochastically lowers the effective threshold for a subset of pixels, generating additional spurious events. Second, we configure a refractory period of $10\,000$~ns ($10~\mu$s) per pixel, bounding the maximum firing rate to
$f_{\max} = \frac{10^{9}}{T_{\text{refrac}}} \;\text{Hz}$
which suppresses high-frequency noise bursts; this value is consistent with the intrinsic refractory period of physical DVS pixels. Third, we increase $\epsilon$ in the log-intensity transform
$L = \log\!\left(\epsilon + \frac{I}{255}\right)$
from 0.001 to 0.01, compressing the dynamic range in the near-dark region and reducing sensitivity to small absolute intensity variations. Together, these modifications reduce the background noise event rate by approximately an order of magnitude while retaining over 95\% of motion-correlated events.

\section{Experiments}

\subsection{Evaluation Metrics}
We adopt standard metrics for both object detection and multi-object tracking to enable fair comparison with existing methods.
Detection quality is measured via precision and recall where $TP$, $FP$, and $FN$ denote true positives, false positives, and false negatives, respectively. A predicted box is considered a true positive if its Intersection over Union (IoU) with a ground-truth box exceeds a given threshold. We report the mean Average Precision (mAP) calculated by averaging the AP values over different IoU thresholds.


We choose the popular MOT evaluation metrics, Multiple Object Tracking Accuracy (MOTA) and Precision (MOTP):
\begin{equation}
\text{MOTA} = 1 - \frac{\sum_{t}(FN_t + FP_t + IDs_t)}{\sum_{t} GT_t},
\end{equation}
\begin{equation}
\text{MOTP} = \frac{\sum_{k=1}^{N}\sum_{i=1}^{M_k} d_{i,k}}{\sum_{k=1}^{N} M_k},
\end{equation}
MOTA captures the combined effect of missed detections, false alarms, and identity switches, while MOTP reflects average localization quality.


\begin{table}[htbp]
\centering
\caption{Object Detection Performance of different Modalities on SEHN Dataset}
\label{tab:detection_evaluation}
\begin{tabular}{llcc}
\toprule
\textbf{Modality} & \textbf{Detector} & \textbf{ALL scenarios} & \textbf{Nighttime-Nolight} \\
 & & mAP (\%)$\uparrow$ & mAP (\%)$\uparrow$ \\
\midrule
RGB & YOLOv11~\cite{khanam2024yolov11} & 83.46 & 46.32 \\
\midrule
Event & RVT~\cite{gehrig2023recurrent} & 65.11 & 63.71 \\
\midrule
RGB+Event & DAGR~\cite{gehrig2024low} & 67.72 & 65.34 \\
\bottomrule
\end{tabular}
\end{table}

\begin{table*}[htbp]
\centering
\caption{Multi-Object Tracking Performance on SEHN Dataset}
\label{tab:dataset_evaluation}
\resizebox{\textwidth}{!}{
\begin{tabular}{llcccccc}
\toprule
\multirow{2}{*}{\textbf{Method}} & \multirow{2}{*}{\textbf{Modality}} & \multicolumn{2}{c}{\textbf{Daytime}} & \multicolumn{2}{c}{\textbf{Nighttime}} & \multicolumn{2}{c}{\textbf{NighttimeNoLight}} \\
\cmidrule(lr){3-4} \cmidrule(lr){5-6} \cmidrule(lr){7-8}
& & MOTA (\%)$\uparrow$ & MOTP (\%)$\uparrow$ & MOTA (\%)$\uparrow$ & MOTP (\%)$\uparrow$ & MOTA (\%)$\uparrow$ & MOTP (\%)$\uparrow$ \\
\midrule
YOLOv11~\cite{khanam2024yolov11} + Bytetrack & RGB & \textbf{87.82} & \textbf{85.48} & 83.26 & 79.56 & 46.44 & 42.10 \\
RVT~\cite{gehrig2023recurrent} + Bytetrack & Event & 78.17 & 74.45 & 70.28 & 65.15 & 68.76 & 65.22 \\
DAGR~\cite{gehrig2024low} + Bytetrack & RGB + Event & 82.55 & 79.89 & 76.90 & 73.26 & 69.27 & 66.55 \\
\midrule
JEAT (RVT+YOLOv11) & RGB + Event & 85.28 & 82.31 & \textbf{84.36} & \textbf{81.29} & \textbf{76.74} & \textbf{71.11} \\
\bottomrule
\end{tabular}
}
\end{table*}

\subsection{Dataset Evaluation}

We first evaluate the object detection performance of the RGB and event modalities separately on the SEHN dataset. We train a standard YOLOv11-Large~\cite{khanam2024yolov11} detector on the RGB frames, an RVT~\cite{gehrig2023recurrent} detector on the event data, and an RGB-Event fusion detector DAGR-M~\cite{gehrig2024low}, all using the same training and validation splits. RVT integrates a MaxViT backbone with convolutional LSTM cells to capture spatiotemporal dependencies in event streams, processing stacked-histogram event representations (10 bins, $\Delta t$=50ms) at $384 \times 640$ resolution across 4 hierarchical stages with channel dimensions [64, 128, 256, 512]. The DAGR-M model is configured with a 50ms event sampling interval and a ResNet-18 as the image backbone. The results are shown in Table~\ref{tab:detection_evaluation}. The RGB-based detector achieves a high mAP of 83.46\%, while the event-based detector RVT achieves a lower mAP of 65.11\%. The RGB-Event fusion detector DAGR improves the mAP to 67.72\%. In the absence of street lighting at nighttime, the performance of the RGB-based detector significantly deteriorates, achieving a mean Average Precision (mAP) of only 46.32\%. In contrast, the event-based detector remains largely unaffected by these challenging illumination conditions, demonstrating robust sensing capabilities in low-light environments.
The RGB-Event fusion method did not achieves a significant improvement overt the event only method, which may be due to the fact that the near completely dark nighttime scenarios in the dataset and the stationary vehicle, where these modality provides very limited information and the fusion method may not effectively leverage the complementary strengths of both modalities.

Table~\ref{tab:dataset_evaluation} summarizes the multi-object tracking performance of our proposed method. Our proposed method with joint fusion significantly outperforms both the RGB-only and event-only baselines, as well as the early fusion method DAGR + Bytetrack. In the daytime scenario, our method achieves a MOTA of 85.28\% and MOTP of 82.31\%, which is lower than the RGB-only baseline due to the additional noise from the event modality. However, in the nighttime scenarios, our method achieves a MOTA of 84.36\% and 76.74\%, which significantly outperforms the RGB-only baseline (83.26\% and 46.44\%) and the event-only baseline (70.28\% and 68.76\%). This demonstrates that our joint fusion method effectively leverages the complementary strengths of both modalities, achieving robust perception in challenging nighttime environments where single-sensor solutions often fail.

Perceiving vehicle trajectories on nighttime highways presents a significant challenge due to the lack of artificial lighting infrastructure. In many scenarios, the absence of streetlights leaves vehicle headlights as the sole source of illumination, which complicates the extraction of reliable visual features. Furthermore, the presence of stationary vehicles on the roadway remains a primary cause of high-speed traffic accidents, as these obstacles are particularly difficult to detect and track under low-visibility conditions. In our experiments, our method leverages the advantages of event-based data in low-light and high-speed scenarios while simultaneously utilizing RGB data to compensate for the inability of event sensors to perceive stationary objects. By integrating these complementary modalities, the proposed method achieves robust perception in challenging environments where single-sensor solutions often fail.


\section{Discussion \& Conclusion}

This paper presented JEAT, a joint adaptive tracking framework that fuses asynchronous RGB and event camera detections through unified data association with online noise adaptation. By replacing fixed-priority sensor selection with a covariance-calibrated cost matrix, JEAT adaptively shifts modality preference under varying illumination and motion conditions. We also introduced SEHN, a synthetic multimodal highway dataset covering daytime, nighttime, and unlit nighttime scenarios with co-registered RGB and event streams, together with a noise mitigation strategy for ultra-low-light DVS simulation. Experiments on SEHN show that JEAT outperforms unimodal and early-fusion baselines in degraded conditions, maintaining robust tracking in completely unlit environments where RGB-only methods fail and event-only methods struggle with stationary vehicles.
A practical advantage of JEAT is its detector-agnostic late-fusion design: RGB and event detectors can be independently selected, trained, or replaced with off-the-shelf models. Since fusion is performed only through lightweight Kalman filtering and linear assignment, the framework adds negligible computational overhead, making it suitable for resource-constrained highway edge deployment. Future work will incorporate explicit camera models to project tracking states into metric world coordinates for more accurate geometry and velocity estimation, and validate the framework on real-world event-camera highway data to bridge the sim-to-real gap.

\section*{Acknowledgment}
This work was supported in part by the National Natural Science Foundation of China (No.72071214), Guangdong  Science and Technology Innovation Foundation (No. 2025B0909020003) and Shenzhen Science and Technology Program under Grant (No. KJZD20240903100802004,ZDCY20250901111359002, ZDCY20250901112001002, GXWD 20231130153844002).

\bibliographystyle{IEEEtran}
\bibliography{ref}

\end{document}